\documentclass{article}

\PassOptionsToPackage{numbers, compress}{natbib}

\usepackage[preprint]{neurips_2025}

\usepackage[utf8]{inputenc} 
\usepackage[T1]{fontenc}    
\usepackage{hyperref}       
\usepackage{url}            
\usepackage{booktabs}       
\usepackage{amsfonts}       
\usepackage{nicefrac}       
\usepackage{microtype}      
\usepackage{xcolor}         
\usepackage{amsmath}
\usepackage{amssymb}
\usepackage{mathtools}
\usepackage{amsthm}
\usepackage{multirow}
\usepackage{adjustbox}
\usepackage{enumitem}
\usepackage{subcaption}
\usepackage[most]{tcolorbox}
\usepackage{listings}
\title{Shuttle Between the Instructions and the Parameters of Large Language Models}

\author{
Wangtao Sun$^{1,2}$, Haotian Xu$^{3}$, Huanxuan Liao$^{1,2}$, Xuanqing Yu$^{1,2}$, \\
\textbf{Zhongtao Jiang}$^{4}$, \textbf{Shizhu He}$^{1,2}$, \textbf{Jun Zhao}$^{1,2}$, \textbf{Kang Liu}$^{1,2,5}$\thanks{Corresponding author: kliu@nlpr.ia.ac.cn} \\
\textit{$^{1}$The Laboratory of Cognition and Decision Intelligence for Complex Systems,} \\
\textit{Institute of Automation, Chinese Academy of Sciences, Beijing, China} \\
\textit{$^{2}$School of Artificial Intelligence, University of Chinese Academy of Sciences, Beijing, China} \\
\textit{$^{3}$Xiaohongshu Inc} \textit{$^{4}$Kuaishou Technology}
\textit{$^{5}$Shanghai Artificial Intelligence Laboratory} \\
}

\begin{document}
\maketitle

\begin{abstract}
The interaction with Large Language Models (LLMs) through instructions has been extensively investigated in the research community. While instructions have been widely used as the guidelines for task solving, this paper further notices that both instructions and parameters are the compression of task data.
Therefore, they could be strongly correlated and can be learned to predict one from the other.
This paper proposes a novel neural network framework, SHIP (\textbf{Sh}uttle between the \textbf{I}nstructions and the \textbf{P}arameters), to model and learn the mutual mappings between the instructions and the parameters of LLMs. 
We verify that SHIP can effectively map one of the instructions/parameters to the other by evaluating it on the tasks of instruction deduction and induction.
The results show that SHIP performs better than existing baseline methods in terms of deductive capabilities while significantly surpassing them in inductive capabilities. 
Moreover, SHIP can effectively combine the two mapping processes to perform excellent inductive reasoning.
The code and data for this paper are released at https://anonymous.4open.science/r/Shuttle-Between-Instructions-Parameters/.
\end{abstract}

\section{Introduction}
\label{sec:intro}

With the rise of Large Language Models (LLMs), an increasing number of researchers and practical applications are beginning to explore interacting with LLMs through \emph{instructions}. Instructions are a type of natural language that delineates task objectives, providing another dimension of supervision for expressing task semantics \cite{yin2023llm}. Unlike instance-level annotations that emphasize specific input-output mappings, instructions encapsulate higher-order task semantics, and are the compression of task data by humans through natural language.

On the other hand, the numerical parameters in LLMs constitute an alternative mechanism for encoding abstract task knowledge. When performing supervised fine-tuning (SFT) on LLMs, the model will learn and compress novel task-specific information through parameter updates. From an information-theoretic perspective, these optimized numerical parameters are the compression of task data in neural networks \cite{deletang2023language}. This formulation naturally suggests that the parameters of an LLM after training on specific tasks should exhibit a high degree of correlation with the task instructions. Therefore, a research question arises: \textbf{could we enable the shuttling (build the mutual mappings) between the instructions and the parameters of LLMs?}. 


Establishing such shuttling offers dual strategic benefits: 1) \emph{From-Instructions-to-Parameters}. Mapping from the instructions to the parameters of LLMs could facilitate rapid model adaptation to novel tasks through parametric adaptation, thereby circumventing extensive SFT data requirements. 2) \emph{From-Parameters-to-Instructions}. 
Mapping from the optimized parameters of LLM to the instructions could enable the verification of learned representations through human-comprehensible explanations, and direct manipulation of model behavior via linguistic instruction editing.

\begin{figure}[htp]
  \centering
  \begin{subfigure}[t]{0.48\textwidth}
    \includegraphics[width=\textwidth, height=5cm, keepaspectratio]{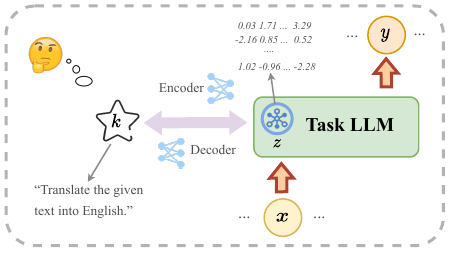}
\caption{The basic concept of learning the mutual mappings between the instructions and the parameters of LLMs.}
\label{fig:concept}
  \end{subfigure}
  \hfill
  \begin{subfigure}[t]{0.48\textwidth} 
    \includegraphics[width=\textwidth, height=5cm, keepaspectratio]{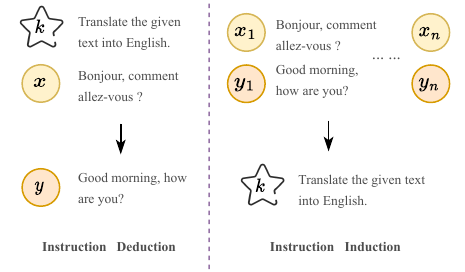}
\caption{The examples of the verification tasks: Instruction Deduction and Instruction Induction.}
\label{fig:task_demonstrations}
  \end{subfigure}
  \caption{The concept of shuttling between the instructions and parameters, and the verification tasks.}
\end{figure}

To this end, this paper proposes SHIP (\textbf{Sh}uttle between the \textbf{I}nstructions and the \textbf{P}arameters), a novel framework to model and learn the mutual mappings between the instructions and the parameters. As illustrated in Figure~\ref{fig:concept}, we denote the textual instruction is as $k$, freeze the main part of the task LLM and take a small part of the trainable parameters as the parameters $z$, while $z$ should be conditioned on the task data $x_i, y_i$. Based on the training data $k,x,y$, an encoder and a decoder will be jointly and end-to-end trained to learn the mutual mappings between the instructions $k$ and the parameters $z$.

To verify the effectiveness of the SHIP framework, we adopt two verification tasks: \emph{instruction deduction} \cite{zhou2023instruction, ivison-etal-2023-hint, liao2024instance} and \emph{instruction induction} \cite{wang2023hypothesis,qiu2023phenomenal,sun2024itd}. As shown in Figure~\ref{fig:concept},\ref{fig:task_demonstrations}:

\begin{itemize}[leftmargin=*]
    \item Given an instruction $k$, task inputs $x$, instruction deduction requires the model to predict the target $y$. In this task, SHIP will map the instruction $k$ to the parameters $z$, and evaluate the deductive inference performance by predicting $y$ (\S\ref{sec:deductive_inference}).
    \item Given multiple instances $(x_i, y_i)$ that satisfy the same instruction, instruction induction demands the model to generate the shared latent instruction $k$. In this task, SHIP will first fine-tune the Task LLM on $(x_i, y_i)$ to obtain the converged parameters $z$, and evaluate the inductive inference performance by predicting $k$ (\S\ref{sec:inductive_inference}).
\end{itemize}


To empirically evaluate the efficacy of SHIP, we conduct a series of experiments and draw the following conclusions:

\textbf{SHIP can effectively map one of the instructions/parameters to the other.} Evaluated on separate deduction and induction tasks, the experimental results revealed that SHIP effectively models and learns the mutual mappings between the instructions and the parameters (Table~\ref{tab:main_results}).
On the task of instruction deduction, SHIP demonstrates better deductive capabilities than SFT and other existing deduction methods. It means that our model could directly and effectively inject the task knowledge in instructions into the parameters of LLMs without acquiring sufficient labeled training examples for SFT. 
On the task of instruction induction, SHIP achieved more than a 40\% improvement in in-distribution performance and over a 20\% improvement in out-of-distribution performance compared to SFT and other existing deduction methods. It denotes that SHIP effectively maps the parameters to instructions and could excellently explain the learned knowledge in LLMs.


\textbf{SHIP can combine the mappings to perform inductive reasoning.} 
\emph{inductive reasoning} requires the model to infer $y$ with an input $x$ and few-shot demonstrations $x_1,y_1;x_2,y_2;...;x_n,y_n$. We find that in this task,  
SHIP can effectively combine the two mappings between instructions and parameters, inducing a $k$ from the few-shot demonstrations and then apply it to $x$ to infer $y$. Compared to baseline methods such as In-Context Learning and Instruction Induction \cite{honovich-etal-2023-instruction}, SHIP is much more effective in conducting inductive reasoning, with a 10\% relative improvement on both seen and unseen tasks (Table~\ref{tab:inductive reasoning}). To analyze the mechanism of SHIP in inductive reasoning, we perform the t-SNE dimensionality reduction on the learned parameters. We observed that the induction-deduction process of SHIP significantly improved the distribution of the learned parameters, thereby enabling the Task LLM to achieve superior reasoning performance (Figure~\ref{fig:sni_latent},\ref{fig:p3_latent}).

In summary, this paper makes the following contributions:
\begin{itemize}[itemsep=1pt,topsep=1pt,parsep=0pt,leftmargin=*]
    \item We propose a novel neural network framework, SHIP, that is designed to model and learn the mutual mappings between the instructions and the parameters. 
    \item We demonstrate that SHIP can effectively map one of the instructions/parameters to the other through instruction deduction/induction tasks. The results show that SHIP performs better than existing baseline methods in terms of deductive capabilities while significantly surpassing them in inductive capabilities.
    \item We further show that SHIP can combine the two mappings between instructions and parameters to achieve superior inductive reasoning performance.
\end{itemize}

\section{SHIP}
\label{sec:SHIP}

\begin{figure*}[ht]
\begin{center}
\centerline{\includegraphics[width=0.9\textwidth]{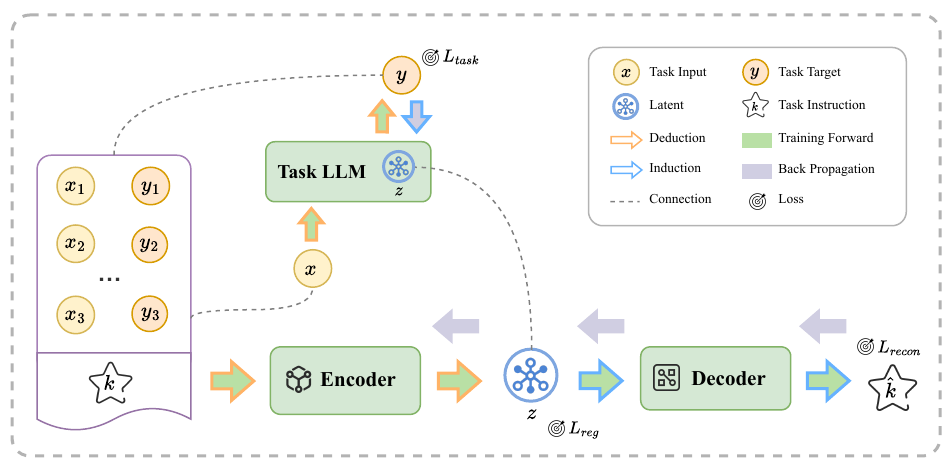}}
\caption{The framework of SHIP. The Training process is represented with filled colors and the inference process is represented with border colors.}
\label{fig:framework}
\end{center}
\end{figure*}

In this section, we introduce the framework, training, and inference of SHIP. SHIP is designed to model and learn the mutual mappings between the instructions $k$ and the corresponding parameters $z$. To achieve this goal, we design a hybrid framework of Variational Autoencoder (VAE \cite{vae}) and Variational Information Bottleneck (VIB \cite{vib}). Specifically, as shown in Figure~\ref{fig:framework}, we ask the latent $z$ not only can serve as the extra parameters of the task LLM to predict the task target $y$ given the input $x$ (VIB), but also to be used to reconstruct the instructions $k$ (VAE). Therefore, SHIP is mainly composed of three models:

\begin{itemize}[itemsep=1pt,topsep=1pt,parsep=0pt,leftmargin=*]
    \item \textbf{Encoder}. A textual encoder that encode the instruction $k$ to the latent $z$. This mapping is denoted as $Enc(\cdot)$.
    \item \textbf{Decoder}. An auto-regressive decoder that decode the latent $z$ to the instructions $k$. The generating distribution of decoder is denoted as $p_{dec}(\cdot)$.
    \item \textbf{Task LLM}. An LLM that solve the downstream task. The generating distribution of Task LLM is denoted as $p_{task}(\cdot)$.
\end{itemize}

In the following part of this section, we will introduce 1) how the three models in SHIP is jointly and end-to-end trained (\S\ref{sec:training}), and 2) how SHIP map the instructions and parameters to each other to solve the instruction deduction (\S\ref{sec:deductive_inference}) and induction (\S\ref{sec:inductive_inference}) tasks.

\subsection{Training}
\label{sec:training}
As shown in Figure~\ref{fig:framework}, the training data for SHIP consists of triples $(k,x,y)$, where $k$ is the instructions, $x, y$ are the input-target pairs that $y$ can be inferred from $x$ using the instructions $k$. Note that there could be multiple $x,y$ that share the same $k$.

First, SHIP uses the Encoder to encode the instruction $k$ into a high-dimension diagonal normal distribution that is parameterized by the mean $\mu$ and covariance $\Sigma$, the latent $z$ is sampled from the encoded normal distribution:
\begin{align}
    \mu, \Sigma = Enc(k), \quad z \sim \mathcal{N}(\cdot|\mu, \Sigma)
\end{align}
Here, the reparametrization trick \cite{reparameterization} is adopted to maintain the gradient flow.
SHIP then uses the Decoder to attempt to reconstruct the instruction $k$ from the latent representation $z$, and calculate the reconstruction loss $L_{recon}$. This corresponds to the objective of VAE.
\begin{align}
    L_{recon} &= -\log p_{dec}(k|z)
\label{eq:original_l_recon}
\end{align}
Meanwhile, the latent representation $z$ is taken as the extra parameters of the Task LLM. The Task LLM is asked to infer on the given task instance $x,y$, and calculate the task loss $L_{task}$. This corresponds to the objective of VIB.
\begin{equation}
\label{eq:original_l_task}
    L_{task} = -\log p_{task}(y|z;x)
\end{equation}
To maintain and leverage the existing well-trained natural language distribution of the Task LLM and auto-regressive Decoder, 
we add textual condition: instruction $k$ for the Task LLM, and one pair of instance $x,y$ for the Decoder (ablation results are in Table~\ref{tab:ablation}). So the Eq~\ref{eq:original_l_recon},\ref{eq:original_l_task} become into:
\begin{align}
    L_{recon} &= -\log p_{dec}(k|z;\underline{x,y}) \\
    L_{task} &= -\log p_{task}(y|z;x,\underline{k})
\end{align}
To prevent the latent $z$ from being overly complex and thus cause overfitting, we calculate the Kullback–Leibler (KL) divergence between $z$ and the standard normal distribution as the regularization loss $L_{reg}$.
\begin{align}
    L_{reg} &= D_{KL}(\mathcal{N}(\cdot|\mu, \Sigma) || \mathcal{N}(\cdot|0, I))
\end{align}
The final objective function is the weighted sum of the three loss terms, and we minimize it under the distribution of training data.
\begin{equation}
    L = \mathbb{E}_{(k,x,y)\sim p_{data}} w_{0} L_{reg} + w_{1} L_{task} + w_2 L_{recon}
\end{equation}
\subsection{Deductive Inference: Map Instructions to Parameters}
\label{sec:deductive_inference}
As indicated by the orange border arrows in the Figure~\ref{fig:framework}, deductive inference is to infer the task output $y$ given the instructions $k$ and the task input $x$. 
To perform deductive inference, SHIP maps the instruction $k$ to the parameters $z$, and uses the Task LLM to generate an output $\hat{y}$ via auto-regressive generation.
\begin{align}
    \mu, \Sigma = & Enc(k), \quad z \sim \mathcal{N}(\cdot|\mu, \Sigma) \\
    \hat{y} &\sim p_{task}(\cdot|z;x,k)
\end{align}
\subsection{Inductive Inference: Map Parameters to Instructions}
\label{sec:inductive_inference}
As indicated by the blue border arrows in the Figure~\ref{fig:framework},
given the multiple instances $T=(x_i,y_i)_{i=1}^n$, inductive inference is to infer their shared instruction $k$.
To infer the instruction $k$, SHIP first fine-tunes the Task LLM on the instances $T$ to obtain the converged latent representation $z^*$ (All other parameters in Task LLMs remain frozen). Instead of directly initializing the $\Tilde{z}$, taking it as the leaf parameters of the computation graph and training it until converge, here, we adopt an \emph{indirect training} trick to fine-tune the Task LLM (The ablation results are in Table~\ref{tab:ablation}). This trick will involve the Encoder into the fine-tuning process and thereby avoids the training-inference inconsistency: 

We first create a trainable tensor $\Tilde{k}$, encode $\Tilde{k}$ into the normal distribution to get the trainable latent $\Tilde{z}$, and compute the task loss: 
\begin{align}
    \Tilde{\mu}, \Tilde{\Sigma} = Enc&(\Tilde{k}), \quad \Tilde{z} \sim \mathcal{N}(\cdot|\Tilde{\mu}, \Tilde{\Sigma}) \\
    J_{task}^{\Tilde{k}}(x,y) &= -\log p_{task}(y|x;\Tilde{z})
\end{align}
Through minimizing $J_{task}^{\Tilde{k}}$ on training task samples $x,y$, we obtain the converged $\Tilde{k}$. With $\Tilde{k}$, we can then obtain the desired converged latent $z^*$:
\begin{align}
    k^* &= \arg\min_{\Tilde{k}} \frac{1}{n} \sum_{i=1}^n J_{task}(x_i,y_i) \\
    \mu^*, \Sigma^* &= Enc(k^*), \quad z^* \sim \mathcal{N}(\cdot|\mu^*, \Sigma^*)
\end{align}
The ablation results of this trick can be found at Table~\ref{tab:ablation}.
After obtaining the parameters $z^*$, we randomly sample a pair of $(x^*,y^*)$ from $T$ to leverage the well-trained natural language distribution of the Decoder. Under this condition, we can decode and map the trained parameters $z^*$ back to explainable instructions $k$:
\begin{align}
    \hat{k} &\sim p_{dec}(\cdot|z^*;x^*, y^*)
\end{align}

\section{Experiment Settings \& Training}
\label{sec:settings}
In the following sections, we conduct a series of experiments to answer the following two research questions:

\begin{itemize}[leftmargin=*]
\item \textbf{RQ1:} Can SHIP learn the mutual mappings between the instructions and the parameters? (\S\ref{sec:exp_induction_deduction})
\item \textbf{RQ2:} Can SHIP combine these mappings together to perform inductive reasoning?  (\S\ref{sec:exp_inductive_reasoning}) 

\end{itemize}

\textbf{Settings.} 
We employ Llama-2-7b-chat \cite{llama2} as the base language model $M$. Task LLM is set to $M$ itself. Encoder, Decoder are $M$ with two LoRA \cite{lora} of rank 16 and 1, respectively.
To facilitate efficient batch training and inference of the Task LLM, we adopt prompt tuning \cite{prompt-tuning} as the trainable parameters $z$ of the Task LLM. The number of soft tokens is set to 10, and thus the dimension of $z$ is $10 \times 4096 = 40960$. All other baselines that need training (later introduced in \S\ref{sec:exp_induction_deduction},\ref{sec:exp_inductive_reasoning}) will take the $z$ of the same size as the training parameters for fair comparisons. The weights of the loss terms $w_0, w_1, w_2$ are set to 1e-3, 1.0, 1.0, respectively.

\begin{figure}[htp]
  \centering
  \begin{subfigure}[t]{0.48\textwidth} 
    \includegraphics[width=\textwidth, height=5cm, keepaspectratio]{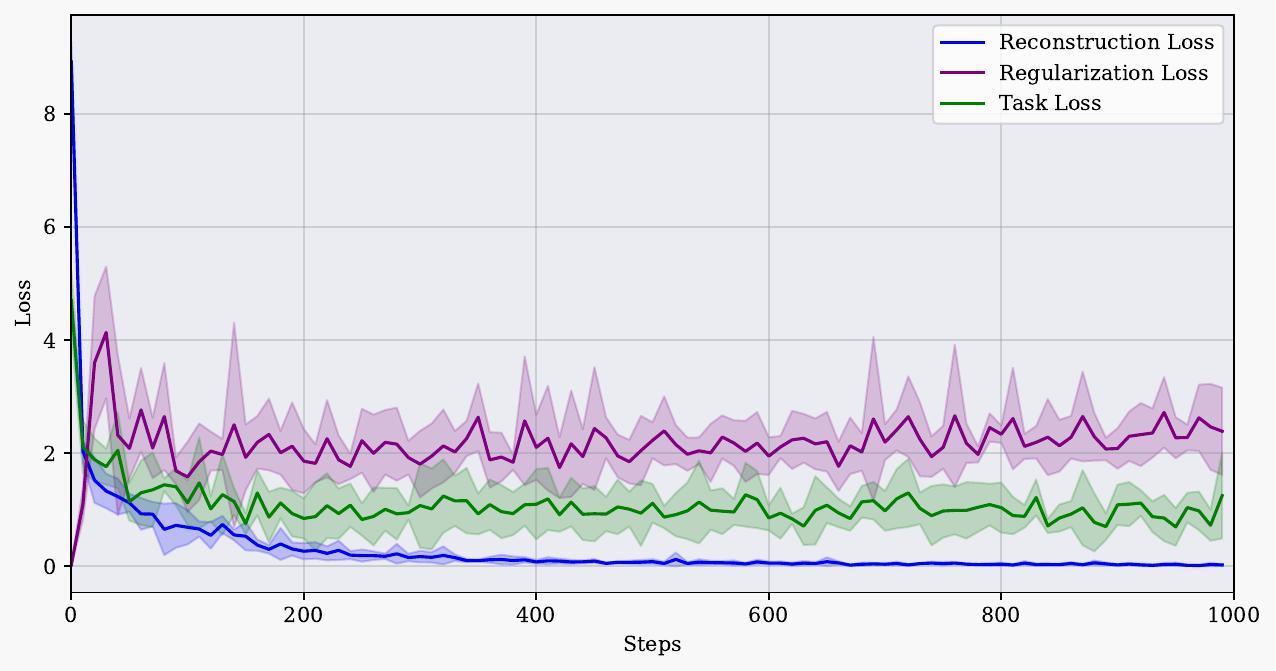}
    \caption{The loss curve of SHIP-in-domain (trained on SNI) with respect to the training steps.}
    \label{fig:loss_domain}
  \end{subfigure}
  \hfill
  \begin{subfigure}[t]{0.48\textwidth}
    \includegraphics[width=\textwidth, height=5cm, keepaspectratio]{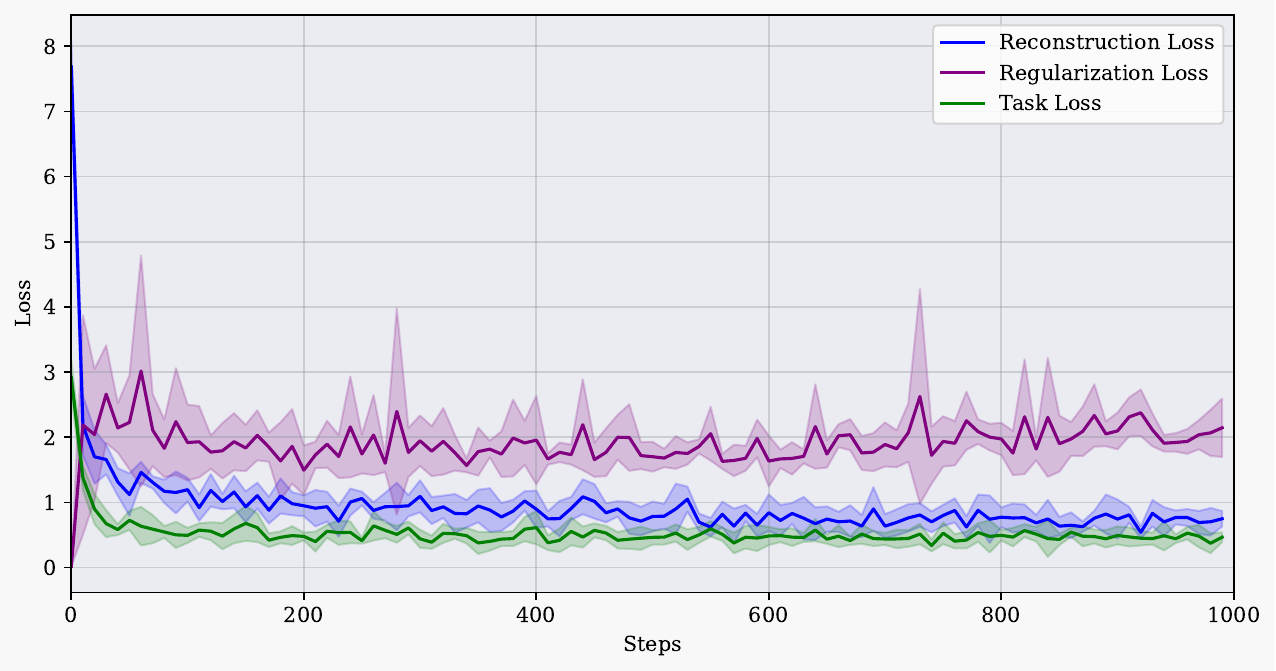}
    \caption{The loss curve of SHIP-pretrain with respect to the training steps.}
    \label{fig:loss_pretrain}
  \end{subfigure}
  \caption{The loss curve of training SHIP.}
\label{fig:loss}
\end{figure}


\textbf{Dataset.} We adopt two popular multi-task instruction datasets: Super-Natural Instructions (SNI, \cite{sni}) and T0 split of P3 (P3, \cite{p3}) for evaluation.
We first split each dataset into seen tasks (90\%) and unseen tasks (10\%).
For each subtask of instruction $k$, we only leave 5 instances $x,y$ as test samples, and use the rest as training samples.
Therefore, for methods that are trained on seen tasks, the test results on seen tasks reflect their \emph{sample-level generalization} ability, while the test results on unseen tasks reflect their \emph{task-level generalization} ability.


\textbf{Training.} Under the above settings, a single run of the experiment is conducted on three NVIDIA A100 GPUs. We trained two versions of SHIP:
\begin{itemize}[itemsep=1pt,topsep=1pt,parsep=0pt,leftmargin=*]
    \item \textbf{SHIP-in-domain}. SHIP that trained on the seen tasks of SNI and P3 for 10 epochs.
    \item \textbf{SHIP-pretrain}. SHIP that trained on around 437k additional instruction following data (Appendix~\ref{app:pretrain_data}) for 1 epoch.
\end{itemize}


In Figure~\ref{fig:loss}, we present the training loss curves for the three terms during the training of both SHIP-in-domain (on the SNI dataset) and SHIP-pretrain, plotted against the training steps. The results are averaged over 6 runs. The initially increased regularization loss (KL term) shows that SHIP is learning and storing increased latent patterns for better reconstruction and task prediction. This is verified in many VAE practices \cite{bowman2015generating}.
The concurrent decrease in both reconstruction loss and task loss demonstrates that SHIP effectively trains the function of the Encoder and Decoder, i.e., learns the mappings between the instructions and the parameters.



\section{SHIP can Effectively Map One of the Instructions/Parameters to the Other}
\label{sec:exp_induction_deduction}
In this section, we verify that SHIP can effectively map one of the instructions/parameters to the other by evaluating SHIP on separate \emph{deduction} and \emph{induction} tasks.
In the deduction task, the model is provided with instructions $k$ and input $x$, and asked to generate the target $y$. In the induction task, the model is provided with 5 test samples $\{x,y\}$ as the observation, and asked to generate the instructions $k$. The examples of deduction and induction are shown in Appendix~\ref{app:examples}.
For the evaluation of both tasks, this paper adopts an external LLM (gpt-4o-mini\footnote{https://openai.com/index/gpt-4o-mini-advancing-cost-efficient-intelligence/}) as a judge to determine whether the prediction is correct. The prompts for the judge and following baseline methods are shown in the Appendix~\ref{app:prompt_judge}.
We adopt the following methods as the baselines:
\begin{itemize}[itemsep=1pt,topsep=1pt,parsep=0pt,leftmargin=*]
    \item \textbf{prompting}. This method simply prompts the LLM $M$ with the instructions $k$ and input $x$ to infer the target $y$ (deduction) 
    and prompts the LLM $M$ with multiple instances $(x,y)$ to infer the instructions $k$ (induction).
    \item \textbf{vanilla SFT}. We fine-tune the LLM $M$ based on the training data of seen tasks to learn the tasks of deduction and induction. The fine-tuned LLM is then evaluated based on the prompting method.
    \item \textbf{TAGI}. TAGI \cite{liao2024instance} is a typical meta-learning-based method that injects the instruction into the Task LLM through the hyper-network. 
    It first trains the ``reference'' parameters of the Task LLM on the training data, and then leverages the (instruction, parameters) pairs to train the hyper-network.
    TAGI can only be used in the \emph{deduction} task.
    \item \textbf{ItD}. ItD \cite{sun2024itd} is a recently proposed method that can empower the induction ability of the language model.
    It first decomposes the joint distribution of $p(x,y,k)$ with a deduction perspective into the instruction prior $p(k)$ and deduction likelihood $p(y|x,k)p(x|k)$, and sample from them.
    Then, it fine-tunes the language model with the sampled data in the form of induction: $p(k|x,y)$.
    ItD can only be used in the \emph{induction} task.
\end{itemize}

\begin{table*}[htp]
\caption{The induction \& deduction performance of SHIP and baselines on SNI and P3. Methods marked with * are not trained on seen tasks. - indicates that the method is not applicable to that task.}
\centering
\begin{adjustbox}{width=\textwidth}
\begin{tabular}{lcccccccc}
\toprule
Dataset & \multicolumn{4}{c}{SNI} & \multicolumn{4}{c}{P3} \\
\midrule
\multirow{2}{*}{method} & \multicolumn{2}{c}{seen tasks (90\%)} & \multicolumn{2}{c}{unseen tasks (10\%)} & \multicolumn{2}{c}{seen tasks (90\%)} & \multicolumn{2}{c}{unseen tasks (10\%)} \\
 & deduction & induction & deduction & induction & deduction & induction & deduction & induction \\
\midrule
prompting * & 12.70 & 20.63 & 12.21 & 26.32 & 21.78 & ~~2.78 & 23.28 & ~~4.76 \\
vanilla SFT & 29.42 & 49.20 & 28.56 & 27.78 & 35.89 & 45.00 & 37.31 & 19.05 \\
TAGI & 32.02  & - & 23.33 & - & 36.33 & - & 47.62 & - \\
ItD & - & 43.85 & - & 33.33 & - & 33.33 & - & \textbf{28.57} \\
\midrule
SHIP-in-domain & \textbf{33.26} & \textbf{85.56} & 21.11 & 44.44 & \textbf{48.67} & \textbf{78.33} & \textbf{58.10} & \textbf{28.57} \\
SHIP-pretrain * & 30.37 & 36.36 & \textbf{32.22} & \textbf{50.00} & 38.22 & 20.00 & 49.52 & 19.05 \\
\bottomrule
\end{tabular}
\end{adjustbox}
\label{tab:main_results}
\end{table*}

\subsection{Comparison with Baselines}
We first compare the accuracy of SHIP on deduction and induction with baselines. As shown in Table~\ref{tab:main_results}, SHIP-in-domain demonstrates better deduction ability compared to SFT and TAGI, and shows impressive induction ability compared to other data-based induction methods, not only outperforming ItD and vanilla SFT on the seen tasks, but also on the unseen tasks by a large margin.
Moreover, the SHIP-pretrain also demonstrates competitive performance on two datasets although it is not trained on the in-domain data.
These results indicate that SHIP effectively maps one of instructions/parameters to the other,  demonstrating excellent 
abilities on both tasks of deduction and induction.

\begin{table*}[htp]
\caption{The ablation results of SHIP on SNI and P3.}
\centering
\begin{adjustbox}{width=\textwidth}
\begin{tabular}{lcccccccc}
\toprule
Dataset & \multicolumn{4}{c}{SNI} & \multicolumn{4}{c}{P3} \\
\midrule
\multirow{2}{*}{method} & \multicolumn{2}{c}{seen task (90\%)} & \multicolumn{2}{c}{unseen task (10\%)} & \multicolumn{2}{c}{seen task (90\%)} & \multicolumn{2}{c}{unseen task (10\%)} \\
 & deduction & induction & deduction & induction & deduction & induction & deduction & induction \\
\midrule
SHIP-in-domain & \textbf{33.26} & \textbf{85.56} & \textbf{21.11} & \textbf{44.44} & \textbf{48.67} & 78.33 & \textbf{58.10} & 28.57 \\
\textcolor{gray}{w/o textual condition $x,y$} & 31.65 & ~~0.53 & 16.67 & ~~0.00 & 45.67 & 11.67 & 54.29 & ~~4.76 \\
\textcolor{gray}{w/o textual condition $k$} & 32.94 & 84.49 & ~~4.44 & 22.22 & 48.22 & \textbf{80.00} & 56.19 & \textbf{33.33}\\
\textcolor{gray}{w/o indirect training} & 29.52 & ~~1.59 & 14.74 & ~~0.00 & 38.00 & ~~7.78 & 49.52 & ~~4.76 \\
\midrule
SHIP-pretrain & \textbf{30.37} & 36.36 & \textbf{32.22} & \textbf{50.00} & 38.22 & \textbf{20.00} & \textbf{49.52} & 19.05 \\
\textcolor{gray}{w/o textual condition $x,y$} & 28.77 & ~~0.53 & 27.78 & ~~0.00 & 37.00 & ~~0.56 & 47.62 & ~~0.00 \\
\textcolor{gray}{w/o textual condition $k$} & 18.29 & \textbf{36.90} & 10.00 & 44.44 & 24.44 & 19.44 & 31.43 & \textbf{28.57} \\
\textcolor{gray}{w/o indirect training} & 28.98 & ~~2.14 & 26.67 & ~~0.00 & \textbf{40.72} & ~~4.42 & 48.57 & ~~3.57 \\
\bottomrule
\end{tabular}
\end{adjustbox}
\label{tab:ablation}
\end{table*}

\subsection{Ablations}
\label{sec:ablation}
To verify the effectiveness of the textual condition (\S\ref{sec:training}) and the indirect training trick (\ref{sec:inductive_inference}), we conduct ablation experiments of SHIP by dropping these parts. 
As shown in Table~\ref{tab:ablation}, if dropping the textual condition $x,y$ for the Decoder, or tuning $z$ without the indirect training trick, the induction performance will greatly decrease. If dropping the textual condition $k$ for the Task LLM, the deduction performance will be harmed to some extent. These findings verify that the textual conditions and indirect training tricks we adopt are essential for the training and inference of SHIP.

\subsection{Generalization with Scaling Up}
\begin{figure}[htp]
  \centering
  \begin{subfigure}[t]{0.48\textwidth} 
    \includegraphics[width=\textwidth, height=5cm, keepaspectratio]{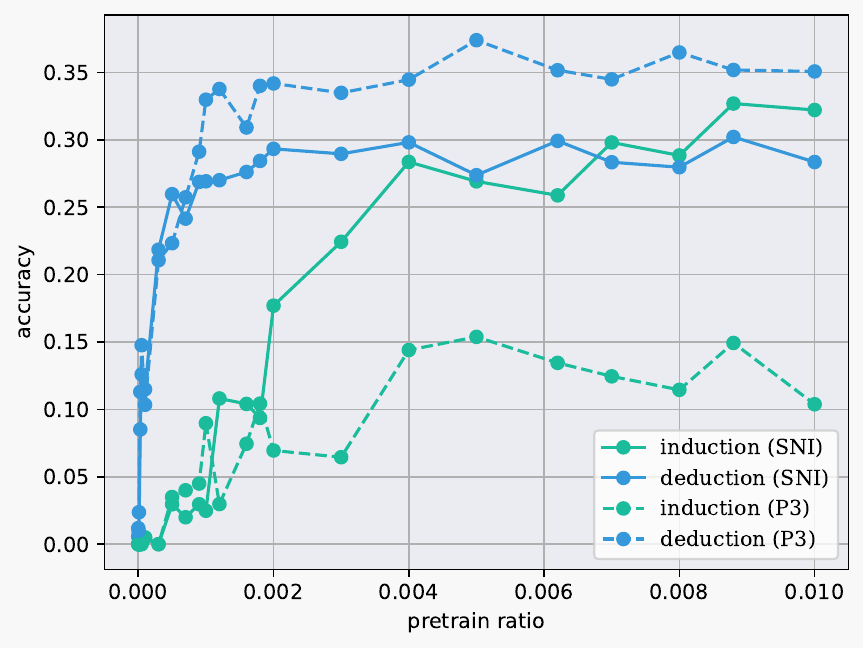}
    \caption{The OOD induction \& deduction performance of SHIP-pretrain with respect to the ratio of used pretrained data.}
    \label{fig:generalization}
  \end{subfigure}
  \hfill
  \begin{subfigure}[t]{0.48\textwidth}
    \includegraphics[width=\textwidth, height=5cm, keepaspectratio]{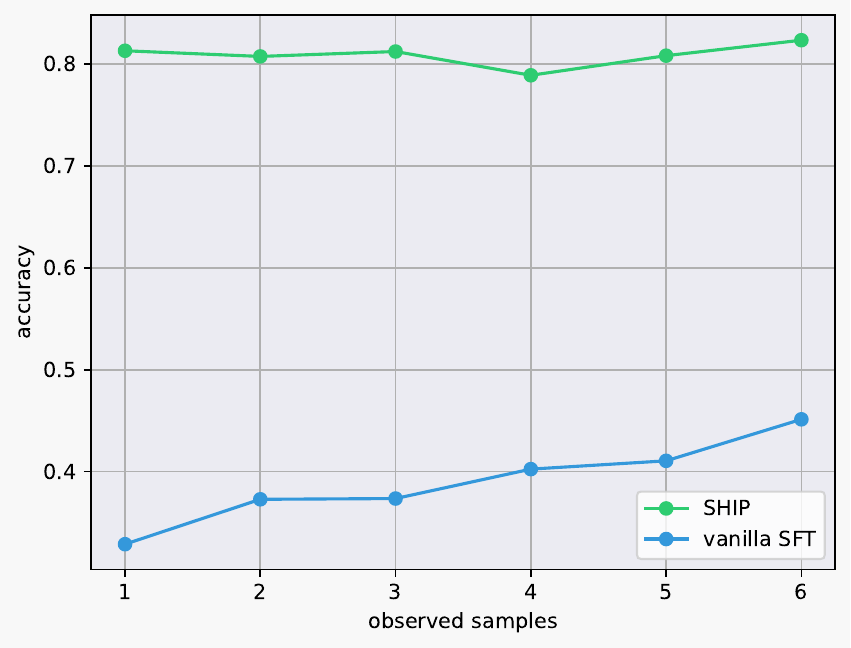}
    \caption{The induction performance of SHIP and SFT on SNI with respect to the number of observed samples. The accuracy is the average accuracy over all seen and unseen tasks.}
    \label{fig:induction}
  \end{subfigure}
  \caption{Analysis of SHIP's generalization ability and few-shot induction ability.}
\end{figure}

To visualize the generalization process of SHIP, we train SHIP-pretrain using varying proportions of the entire pretraining dataset and evaluate its performance on the induction and deduction tasks for SNI and P3. The resulting performance curve is depicted in Figure~\ref{fig:generalization}.
From the curve, it is evident that SHIP's induction and deduction capabilities improve progressively as the volume of pretraining data increases. Notably, the deduction ability exhibits rapid growth and early convergence with increasing pretraining data, whereas the induction ability converges at a later stage. This observation aligns with the perspective highlighted in prior works \cite{bang2023multitask, semanticThanSymbolic, sun2024itd}, which posit that ``induction is harder than deduction for LLMs.''

\subsection{Few-shot Induction}
To further highlight the superiority of SHIP, we conduct a comparative analysis between SHIP and vanilla SFT across varying numbers of observed samples. Specifically, we train vanilla SFT with 1 to 6 observed samples and evaluate both SHIP and vanilla SFT using the corresponding number of testing observations.
As illustrated in Figure~\ref{fig:induction}, SHIP achieves nearly optimal induction performance even when observing just 1 sample, whereas vanilla SFT requires more observed samples to gradually improve its induction capability. These results reveal SHIP's superiority in few-shot induction, demonstrating its ability to perform effectively even in one-shot induction scenarios.


\section{SHIP can Combine the Mappings to Perform Inductive Reasoning}
\label{sec:exp_inductive_reasoning}
To verify whether SHIP can effectively 
combine the two mappings between instructions and parameters
, we further consider the \emph{inductive reasoning} task.
In this task, models are asked to infer $y$ with an input $x$ and few-shot demonstrations $x_1,y_1;x_2,y_2;...;x_n,y_n$. 
Inductive reasoning demands the model to combine the abilities of inductive reasoning and the deductive reasoning: the model is first supposed to \emph{induce} the latent instruction $k$ from the given observations $x_i,y_i$, and then apply it to the test input $x$ to \emph{deduce} the prediction $y$. We adopt the following inductive reasoning methods for comparison:
\begin{itemize}[itemsep=1pt,topsep=1pt,parsep=0pt,leftmargin=*]
    \item \textbf{ICL}. We adopt in-context learning (ICL) as the basic method of inductive reasoning. Specifically, we splice the observations $x_i, y_i$ and the input $x$ together into a prompt: $x_1,y_1;x_2,y_2;...;x_n,y_n;x$, and let the LLM to generate the correspond $y$.
    \item \textbf{Instruction Induction}. Instruction Induction \cite{honovich-etal-2023-instruction} proposed to explicitly induce textual instruction $k$ from the observations $x_1,y_1;x_2,y_2;...;x_n,y_n$, and then prompt the LLM with the query $x$ and instruction $k$ to perform inductive reasoning.
    \item \textbf{SHIP-SFT}. With the well-trained SHIP, First, follow the inference process in \S\ref{sec:inductive_inference}, we fine-tune the Task LLM on the demonstrations, to obtain the converged parameters $z^*$. We use this fine-tuned $z^*$ for the Inductive Inference $p_{task}(\cdot|z^*;x)$ (\S\ref{sec:inductive_inference}).
    \item \textbf{SHIP-Refined}. In this method, we combine the two mappings between instructions and parameters of SHIP to perform inductive reasoning. We first leverage the $z^*$ to decode the induced instructions $\hat{k}$. Then, follow the inference process in \S\ref{sec:deductive_inference}, we again encoded the instruction $\hat{k}$ into $\hat{z}$, and finally infer $y$ with $p_{task}(\cdot|\hat{z};x)$. Note that although we have obtained the instructions $k$ and we have proved it beneficial for deduction \S\ref{sec:ablation}, we do not add it as the additional textual condition (i.e. $p_{task}(\cdot|\hat{z};x,k)$) as we want to directly compare the quality of $z$. 
\end{itemize}

\begin{table*}[htp]
\caption{The inductive reasoning results of SHIP and baselines on SNI and P3.}
\centering
\begin{adjustbox}{width=\textwidth}
\begin{tabular}{llcccc}
\toprule
Dataset & & \multicolumn{2}{c}{SNI} & \multicolumn{2}{c}{P3} \\
\midrule
method & & seen task (90\%) & unseen task (10\%) & seen task (90\%) & unseen task (10\%) \\
\midrule
ICL & & 10.91 & 14.44 & 13.22 & 22.86 \\
Instruction Induction & & 12.80 & ~~7.37 & 17.00 & 27.62 \\
\midrule
\multirow{2}{*}{SHIP-in-domain} & SFT       & 11.98 & ~~5.56  & 27.00 & 30.48 \\
                                      & Refined   & \textbf{33.33} & 16.11  & \textbf{46.89} & \textbf{59.05} \\
\multirow{2}{*}{SHIP-pretrain} & SFT       & ~~3.42 & ~~3.33  & 10.89  & 21.90  \\
                                      & Refined   & 21.39 & \textbf{20.00} & 26.56 & 33.33 \\
\bottomrule
\end{tabular}
\end{adjustbox}
\label{tab:inductive reasoning}
\end{table*}

\subsection{Comparison with Baselines}

As illustrated in Table~\ref{tab:inductive reasoning}, the direct fine-tuned $z^*$ (SHIP-SFT) demonstrates limited effectiveness in assisting the Task LLM to predict $y$ based on $x$. However, a significant improvement is observed when $z^*$ is first decoded into $\hat{k}$ and subsequently re-encoded into $\hat{z}$. This approach substantially enhances the Task LLM's performance with $\hat{z}$, surpassing the ICL baseline by a considerable margin. These findings suggest that SHIP effectively integrates its \emph{deductive} and \emph{inductive} capabilities to facilitate \emph{inductive reasoning}.


\subsection{Semantic Distribution of the Latent}

\begin{figure}[htp]
  \centering
  \begin{subfigure}[t]{0.48\textwidth} 
    \includegraphics[width=\textwidth, height=5cm, keepaspectratio]{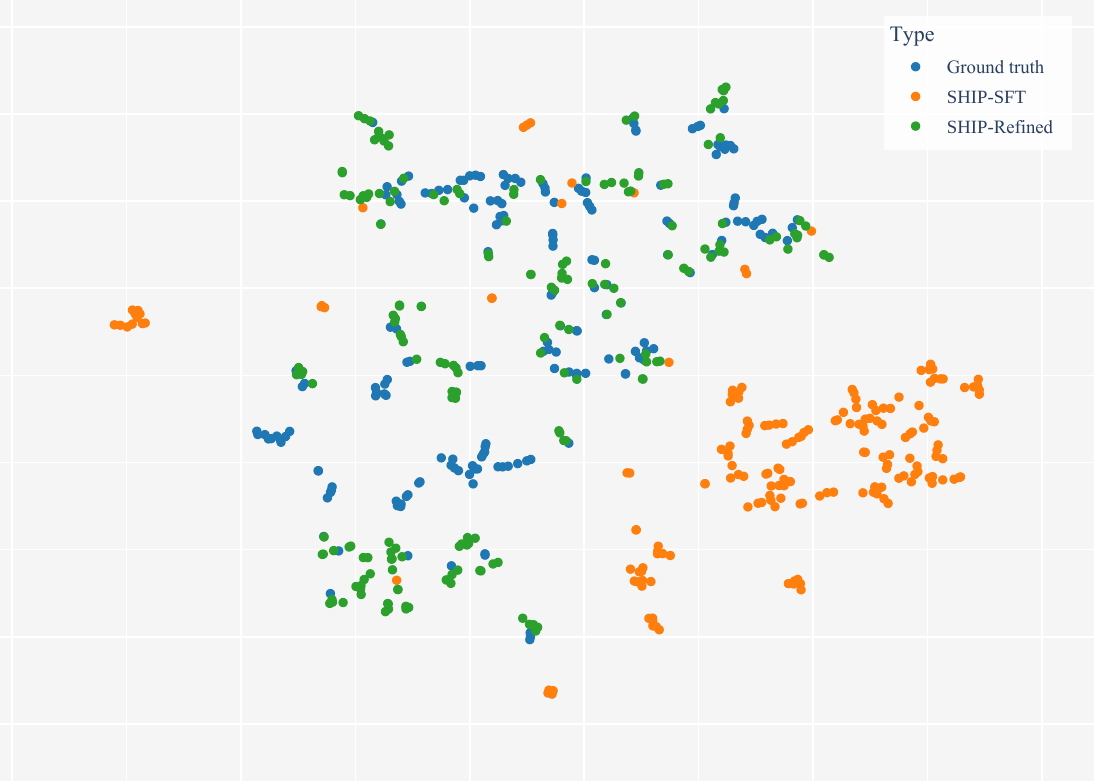}
    \caption{The t-SNE result of latent $z$ on SNI.}
    \label{fig:sni_latent}
  \end{subfigure}
  \hfill
  \begin{subfigure}[t]{0.48\textwidth}
    \includegraphics[width=\textwidth, height=5cm, keepaspectratio]{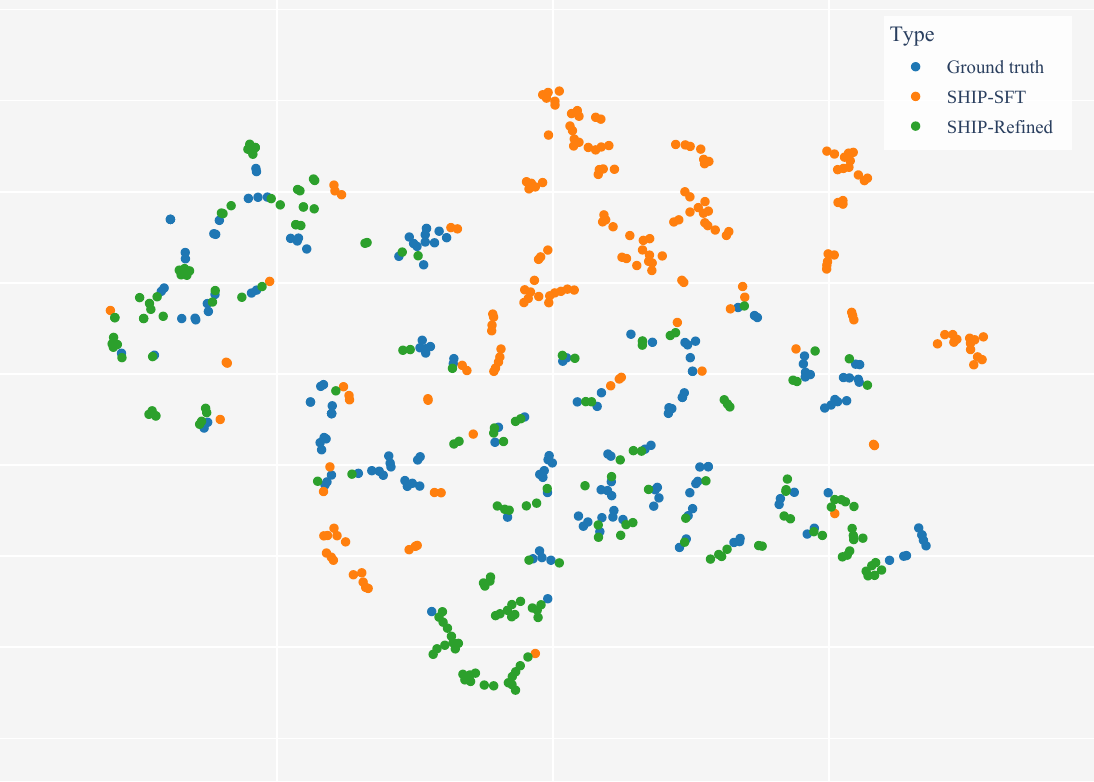}
    \caption{The t-SNE result of latent $z$ on P3.}
    \label{fig:p3_latent}
  \end{subfigure}
  \caption{The t-SNE result of latent $z$.}
\end{figure}

To elucidate why the decode-encode collaborative process of $z$ significantly enhances SHIP's inductive reasoning capabilities, we generate and analyze three distinct types of $z$:
\begin{itemize}[itemsep=1pt,topsep=1pt,parsep=0pt,leftmargin=*]
    \item \textbf{Ground truth}. We use the SHIP to encode the annotated $k$ of the dataset into $z$.
    \item \textbf{SHIP-SFT}. The trained $z^*$ after \textbf{SHIP-SFT}.
    \item \textbf{SHIP-Refined}. The $\hat{z}$ that is obtained by \textbf{SHIP-Refined}.
\end{itemize}
We employ t-SNE \cite{tsne} for dimensionality reduction, projecting all $z$ into a 2D plane and differentiating their types with distinct colors. As depicted in Figure~\ref{fig:sni_latent} and Figure~\ref{fig:p3_latent}, the trained latent from SFT significantly deviates from the ground truth latent. However, by performing the induction-deduction collaborative process, the refined latent becomes markedly closer to and aligned with the ground truth (green and blue). These findings demonstrate that SHIP effectively refines the trained latent, adapting it to better align with true semantic representations, thereby enhancing its inductive reasoning performance.

\section{Related Work}
\label{sec:related_work}
\textbf{Instruction-based LLM Deduction.} Given the task instruction and an input, how to enable LLM to faithfully perform deduction based on it, i.e., instruction following, has been widely considered by researchers. Previous studies, such as IFEval \cite{instruction-following-eval}, InfoBench \cite{qin2024infobench}, and RuleBench \cite{sun2024beyond}, have been instrumental in evaluating the capacity of large models to follow the instructions, also demonstrating that instruction fine-tuning (IFT) can significantly bolster this capability. 
Different from the prompt-level instruction-following paradigm, Meta-Learning methods like Hint \cite{ivison-etal-2023-hint} and TAGI \cite{liao2024instance} have tried training a hyper-network to encode the instruction into some extra parameters of LLMs to execute the instruction. However, these Meta-Learning methods rely heavily on supervised training conducted in advance on each subtask to obtain (instruction, parameter) pairs as training data for the hyper-network.
SHIP employs a similar hyper-network architecture that maps instructions to LLMs' parameters, but it further integrates a reconstruction process, enabling the training of this hyper-network to no longer depend on pre-prepared (parameter, instruction) pairs. Instead, it can be trained on general instruction-following datasets.

\textbf{Instruction-oriented LLM Induction.} For the sake of interpretability and generalization, some previous works also try to induce instruction from task observations through LLMs. Some evaluation studies \cite{eval1, eval2, mitchell2023comparing} have consistently demonstrated that current LLMs are poor at the task of induction. To improve LLMs' capability of induction, methods such as Hypothesis Search \cite{wang2023hypothesis}, Hypothesis Refinement \cite{qiu2023phenomenal}, and ItD \cite{sun2024itd} have modeled induction as a sentence generation task, attempting to enhance the inductive abilities of large models through approaches like sampling-selecting and augmenting-finetuning.
However, these methods are confined to \emph{data-based induction} and overlook the fact that the parameters of neural networks, once trained to converge on task data, provide highly indicative cues for the objectives of induction. SHIP introduces \emph{parameter-based induction}, and our experiments have demonstrated that this approach significantly outperforms the previous series of data-based induction methods.




\section{Conclusion}
\label{sec:conclusion}
\textbf{Contributions.} This paper proposes SHIP, a novel neural network framework that is designed to model and learn the mutual mappings between the instruction and the parameters of the LLMs. A series of experiments is conducted to verify the effectiveness of SHIP, which performs better than existing baseline methods in terms of deductive capabilities while significantly surpassing them in inductive capabilities. Moreover, by combining the process of induction and deduction in SHIP, we find that SHIP can perform excellent inductive reasoning when there is no instruction.


\textbf{Limitations and Outlook.} The scope of deduction and induction is limited to \emph{instruction} in this work, while other forms of task information, such as \emph{rules} may compress more difficult and informative instructions. We will expand and scale up SHIP to this scope in the future.

\bibliography{custom}  
\bibliographystyle{plain} 

\appendix
\newpage


\section{Instruction-following Data for Pretraining SHIP}
\label{app:pretrain_data}
We collect and process the instruction-following data from the following HuggingFace datasets for the pretraining of SHIP:
\begin{itemize}[itemsep=1pt,topsep=1pt,parsep=0pt]
    \item \texttt{xzuyn/manythings-translations-alpaca}
    \item \texttt{MBZUAI/LaMini-instruction}
    \item \texttt{tatsu-lab/alpaca}
    \item \texttt{silk-road/alpaca-data-gpt4-chinese}
    \item \texttt{yizhongw/self\_instruct}
\end{itemize}

\section{Prompts for the LLM Judge and Baselines}
\label{app:prompt_judge}

\begin{figure*}[h]
\begin{tcolorbox}[colframe=blue!50!black, colback=blue!2, title=Prompt for the LLM Judge in Deduction]
\textbf{Role: System} \\
Here are an instruction, an input, an reference answer and a predicted answer. Please help me determine if the predicted answer is correct.
Only return ``True" or ``False". \\
\textbf{Role: User} \\
instruction: \{$k$\} \\
input: \{$x$\} \\
reference answer: \{$y$\} \\
predicted answer: \{$\hat{y}$\}
\end{tcolorbox}
\caption{The prompt for the external LLM to judge if the deduction result $\hat{y}$ is correct for the current case. $k, x, y$ stands for the instruction, the input, and the target answer of the current cases, respectively.}
\label{fig:prompt_deduction}
\end{figure*}

\begin{figure*}[h]
\begin{tcolorbox}[colframe=blue!50!black, colback=blue!2, title=Prompt for the LLM Judge in Induction]
\textbf{Role: System} \\
Here are two instructions described in natural language. 
Please help me determine if these two instructions are equivalent.
Only return ``True" or ``False".  \\
\textbf{Role: User} \\
transformation A: \{$k$\} \\
transformation B: \{$\hat{k}$\}
\end{tcolorbox}
\caption{The prompt for the external LLM to judge if the induction result $\hat{k}$ is correct for the current case. $k$ stands for the instruction of the current cases.}
\label{fig:prompt_induction}
\end{figure*}

\begin{figure*}[h]
\begin{tcolorbox}[colframe=blue!50!black, colback=blue!2, title=Prompt for the Baselines in Deduction]
\textbf{Role: System} \\
\{$k$\} \\
\textbf{Role: User} \\
\{$x$\}
\end{tcolorbox}
\caption{The prompt for the Baselines in Deduction.}
\label{fig:prompt_deduction}
\end{figure*}

\begin{figure*}[h]
\begin{tcolorbox}[colframe=blue!50!black, colback=blue!2, title=Prompt for the Baselines in Induction]
\textbf{Role: System} \\
I gave a friend an instruction and an input. The friend read the instruction and wrote an output for the input. \\
\textbf{Role: User} \\
Here is the input-output pair: \\
Input: \{$x_1$\} \\
Output: \{$y_1$\} \\
... \\
Input: \{$x_5$\} \\
Output: \{$y_5$\} \\
The instruction was:
\end{tcolorbox}
\caption{The prompt for the Baselines in Induction.}
\label{fig:prompt_induction}
\end{figure*}

\section{Examples of Deduction and Induction}
\label{app:examples}
\begin{figure*}[h]
\begin{tcolorbox}[colframe=blue!50!black, colback=blue!2, title=Example for Deduction]
\textbf{Instruction $k$:} \\
In this task, you are given a country name and you need to return the capital city of the given country \\
\\
\textbf{Case 1:} \\
Input $x$: Angola \\
Prediction $\hat{y}$: The capital city of Angola is Luanda. \\
Target $y$: Luanda \\
Score: 1 \\
\\
\textbf{Case 2:} \\
Input $x$: Australia \\
Prediction $\hat{y}$: The capital city of Australia is Sydney. \\
Target $y$: Canberra \\
Score: 0 \\
\\
\textbf{Case 3:} \\
Input $x$: Bangladesh \\
Prediction $\hat{y}$: New Delhi is the capital city of Bangladesh. \\
Target $y$: Dhaka \\
Score: 0 \\
\\
\textbf{Case 4:} \\
Input $x$: Armenia \\
Prediction $\hat{y}$: Yerevan. \\
Target $y$: Yerevan \\
Score: 1 \\
\\
\textbf{Case 5:} \\
Input $x$: Barbados \\
Prediction $\hat{y}$: The capital city of Barbados is Bridgetown. \\
Target $y$: Bridgetown \\
Score: 1
\end{tcolorbox}
\caption{An example case of the deduction task.}
\label{fig:example_deduction}
\end{figure*}

\begin{figure*}[h]
\begin{tcolorbox}[colframe=blue!50!black, colback=blue!2, title=Example for Induction]
\textbf{Observations:} \\
Input $x$: Angola \\
Target $y$: Luanda \\
\\
Input $x$: Australia \\
Target $y$: Canberra \\
\\
Input $x$: Bangladesh \\
Target $y$: Dhaka \\
\\
Input $x$: Armenia \\
Target $y$: Yerevan \\
\\
Input $x$: Barbados \\
Target $y$: Bridgetown \\
\\
\textbf{Ground truth instruction $k$:} \\
In this task, you are given a country name and you need to return the capital city of the given country \\
\\
\textbf{Predicted instruction $\hat{k}$:} \\
Name the capital of the given country. \\
\\
\textbf{Score:} \\
1
\end{tcolorbox}
\caption{An example case of the induction task.}
\label{fig:example_induction}
\end{figure*}

\clearpage
\section*{NeurIPS Paper Checklist}

\begin{enumerate}

\item {\bf Claims}
    \item[] Question: Do the main claims made in the abstract and introduction accurately reflect the paper's contributions and scope?
    \item[] Answer: \answerYes{} 
    \item[] Justification: We clearly claim the contributions and scope of this paper in the abstract and introduction. 

\item {\bf Limitations}
    \item[] Question: Does the paper discuss the limitations of the work performed by the authors?
    \item[] Answer: \answerYes{} 
    \item[] Justification: We discuss the limitations of the work in the Section~\ref{sec:conclusion}.

\item {\bf Theory assumptions and proofs}
    \item[] Question: For each theoretical result, does the paper provide the full set of assumptions and a complete (and correct) proof?
    \item[] Answer: \answerNA{} 
    \item[] Justification: This paper does not include theoretical results.

    \item {\bf Experimental result reproducibility}
    \item[] Question: Does the paper fully disclose all the information needed to reproduce the main experimental results of the paper to the extent that it affects the main claims and/or conclusions of the paper (regardless of whether the code and data are provided or not)?
    \item[] Answer: \answerYes{} 
    \item[] Justification: The information needed to reproduce our experimental results is included in the Section~\ref{sec:settings}. 

\item {\bf Open access to data and code}
    \item[] Question: Does the paper provide open access to the data and code, with sufficient instructions to faithfully reproduce the main experimental results, as described in supplemental material?
    \item[] Answer: \answerYes{} 
    \item[] Justification: The data and code can be accessed through the link in the Abstract. 

\item {\bf Experimental setting/details}
    \item[] Question: Does the paper specify all the training and test details (e.g., data splits, hyperparameters, how they were chosen, type of optimizer, etc.) necessary to understand the results?
    \item[] Answer: \answerYes{} 
    \item[] Justification: The settings of experiments are included in the Section~\ref{sec:settings}. 

\item {\bf Experiment statistical significance}
    \item[] Question: Does the paper report error bars suitably and correctly defined or other appropriate information about the statistical significance of the experiments?
    \item[] Answer: \answerNo{} 
    \item[] Justification: The training and evaluation rely on locally deployed and API-accessed LLMs respectively, making repeated experiments computationally expensive.

\item {\bf Experiments compute resources}
    \item[] Question: For each experiment, does the paper provide sufficient information on the computer resources (type of compute workers, memory, time of execution) needed to reproduce the experiments?
    \item[] Answer: \answerYes{} 
    \item[] Justification: We include these details in the Section~\ref{sec:settings}.
    
\item {\bf Code of ethics}
    \item[] Question: Does the research conducted in the paper conform, in every respect, with the NeurIPS Code of Ethics \url{https://neurips.cc/public/EthicsGuidelines}?
    \item[] Answer: \answerYes{} 
    \item[] Justification: The research conducted in the paper conforms, in every respect, with the NeurIPS Code of Ethics.

\item {\bf Broader impacts}
    \item[] Question: Does the paper discuss both potential positive societal impacts and negative societal impacts of the work performed?
    \item[] Answer: \answerNA{} 
    \item[] Justification: We do not foresee any social impact of our work in its current form. We believe its impact should be confined to the academic domain.
    
\item {\bf Safeguards}
    \item[] Question: Does the paper describe safeguards that have been put in place for responsible release of data or models that have a high risk for misuse (e.g., pretrained language models, image generators, or scraped datasets)?
    \item[] Answer: \answerNA{} 
    \item[] Justification: Our models have no such potential for misuse. 

\item {\bf Licenses for existing assets}
    \item[] Question: Are the creators or original owners of assets (e.g., code, data, models), used in the paper, properly credited and are the license and terms of use explicitly mentioned and properly respected?
    \item[] Answer: \answerYes{} 
    \item[] Justification: All code, data, and models used in this paper are properly credited. 

\item {\bf New assets}
    \item[] Question: Are new assets introduced in the paper well documented and is the documentation provided alongside the assets?
    \item[] Answer: \answerYes{} 
    \item[] Justification: We provide a README in our code release, which we plan to gradually improve in our open-source repository.

\item {\bf Crowdsourcing and research with human subjects}
    \item[] Question: For crowdsourcing experiments and research with human subjects, does the paper include the full text of instructions given to participants and screenshots, if applicable, as well as details about compensation (if any)? 
    \item[] Answer: \answerNA{} 
    \item[] Justification: This paper does not involve crowdsourcing nor research with human subjects.

\item {\bf Institutional review board (IRB) approvals or equivalent for research with human subjects}
    \item[] Question: Does the paper describe potential risks incurred by study participants, whether such risks were disclosed to the subjects, and whether Institutional Review Board (IRB) approvals (or an equivalent approval/review based on the requirements of your country or institution) were obtained?
    \item[] Answer: \answerNA{} 
    \item[] Justification: This paper does not involve crowdsourcing nor research with human subjects.

\item {\bf Declaration of LLM usage}
    \item[] Question: Does the paper describe the usage of LLMs if it is an important, original, or non-standard component of the core methods in this research? Note that if the LLM is used only for writing, editing, or formatting purposes and does not impact the core methodology, scientific rigorousness, or originality of the research, declaration is not required.
    \item[] Answer: \answerNA{} 
    \item[] Justification: The core method development in this research does not involve LLMs as any important, original, or non-standard components.

\end{enumerate}

\end{document}